# KGS-GCN: Kinematics-Driven Gaussian Splatting and Probabilistic Topology for Skeleton-Based Action Recognition

Yuhan Chen, Yicui Shi, Guofa Li *Senior Member, IEEE*, Liping Zhang, Jie Li, Jiaxin Gao, Wenbo Chu *Senior Member, IEEE*

***Abstract*—Skeleton-based action recognition is widely applied in sensor-based systems, including human-computer interaction and intelligent surveillance. However, typical sensors produce sparse and discrete joint coordinates, which often results in the loss of fine-grained spatiotemporal information during highly dynamic movements. Furthermore, predefined physical topologies impose significant restrictions on modeling potential long-range dependencies. To address these challenges, KGS-GCN is proposed, which integrates kinematics-driven Gaussian splatting and probabilistic topology within a graph convolutional network. Specifically, a kinematics-driven Gaussian splatting module is designed to dynamically construct anisotropic covariance matrices by extracting instantaneous velocity vectors of joints. This process renders sparse skeleton sequences into multi-view continuous heatmaps rich in spatiotemporal semantics, facilitating an explicit visual representation of motion states. Furthermore, a probabilistic topology construction strategy is introduced to transcend the limitations of physical connectivity. This strategy utilizes the Bhattacharyya distance to quantify statistical correlations between joint Gaussian distributions, thereby generating an adaptive prior adjacency matrix with explicit statistical significance. Finally, the lightweight multi-view rendering branch and the topological GCN backbone are unified through a visual context gating mechanism. This integration enables the pipeline to seamlessly fuse continuous dynamic cues with structural priors while maintaining high computational efficiency, requiring only 1.4M parameters and 1.3 GFLOPs. Extensive experiments on multiple benchmark datasets demonstrate that KGS-GCN significantly enhances the modeling of complex spatiotemporal dynamics and achieves competitive performance at a low computational cost. This framework establishes an efficient paradigm for improving the perceptual robustness of low-fidelity sensor data.**

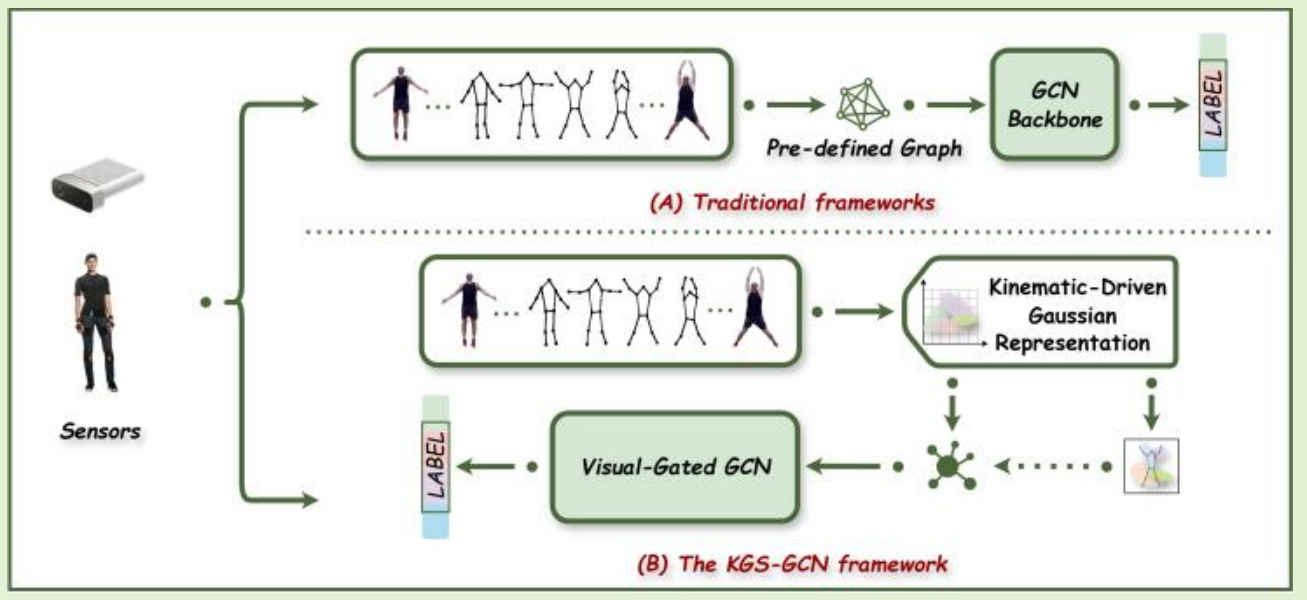

*Index Terms*—*Skeleton-based action recognition; Gaussian splatting; Probabilistic topology learning*

## I. INTRODUCTION

RAPID advancements in micro-electro-mechanical systems and 3D sensing technologies have paved the way for motion-capture-based perception in critical domains, such as intelligent medical rehabilitation, human-robot collaboration, and pervasive computing [1-3]. In these contexts, skeleton data serves as a high-level sensor signal extracted from depth cameras, radars, or inertial measurement units. It can effectively encapsulate human biomechanical structures with minimal bandwidth requirements while ensuring superior privacy protection compared to raw RGB videos. Consequently, these merits have established skeleton data as a fundamental modality for action recognition tasks [2-4].

Traditional sensor signal processing methods rely on manual feature engineering or RNN/CNN-based time-series analysis. However, these approaches struggle to accommodate the non-Euclidean topological structures inherent in human skeletons. To address this, ST-GCN [5] pioneered a strategy to formulate skeleton sensor data as spatiotemporal graphs, utilizing graph convolutional networks to aggregate signals along physical limb connections. This paradigm effectively establishes structured dependencies among sensor nodes, establishing graph neural networks as the dominant architecture for skeleton-based perception tasks.

Subsequent works have focused on overcoming the limitations of fixed physical topology by exploring data-driven topology learning and spatiotemporal interaction mechanisms.

This work was jointly supported by the National Key R&D Program of China under Grant 2024YFB2505500, the National Natural Science Foundation of China under Grant 52272421 and 52372377, and the Young Beijing Scholars Program under Grant 2024-069. *(Corresponding author: Wenbo Chu)*

Yuhan Chen, Yicui Shi, Guofa Li and Jie Li are with the College of Mechanical and Vehicle Engineering, Chongqing University, Chongqing 400044, China (e-mail: 20240701028@stu.cqu.edu.cn; yicuishi@cqu.edu.cn; liguofa@cqu.edu.cn; jieli@cqu.edu.cn).

Liping Zhang is with the Department of Mathematical Sciences, Tsinghua University, Beijing 100084, China (e-mail: lipingzhang@tsinghua.edu.cn).

Jiaxin Gao is with the School of Vehicle and Mobility, Tsinghua University, Beijing 100084, China (e-mail: gaojiaxin2017@163.com).

Wenbo Chu is with the Western China Science City Innovation Center of Intelligent and Connected Vehicles (Chongqing) Co., Ltd., Chongqing 401329, China; the National Innovation Center of Intelligent and Connected Vehicles, Beijing 100176, China; the Engineering Research Center of Mechanical Testing Tech. and Equip. Ministry of Education, Chongqing University of Technology, Chongqing 400054, China (e-mail: chuwenbo@wicv.cn).

For instance, 2s-AGCN [6] introduced adaptive graph convolution to learn data-specific adjacency matrices end-to-end, significantly enhancing feature discriminability. Similarly, Shift-GCN [10] employed shift operations to reduce computational complexity while extending spatiotemporal receptive fields. DeGCN [20] proposed deformable graph convolution to dynamically adjust neighborhood aggregation ranges, accommodating deformation variations across distinct actions. Meanwhile, Transformer-based architectures have leveraged self-attention mechanisms to capture global long-range dependencies, continuously pushing the performance boundaries of this task [1, 21].

Despite these advancements, current skeleton perception methods face two fundamental bottlenecks in sampling and modeling complex motion signals. First, discrete point sampling fails to adequately represent continuous motion signals. Conventional pipelines typically treat joints as isolated points and learn features directly from their coordinate sequences. While this approach remains effective for slow or predictable movements, it fails to capture the intricate dynamics of complex actions. For rapid or explosive movements, instantaneous joint velocities, directions, and momentum serve as critical discriminative factors. Being restricted to sparse coordinates attenuates spatial expansion cues along the motion direction. Consequently, the model becomes prone to confusing actions that exhibit similar trajectories but possess distinct dynamic characteristics. Second, current sensor topology construction lacks statistical interpretability. Although adaptive graph learning can transcend physical connectivity constraints to incorporate latent long-range dependencies [6], the edge weights are typically determined through implicit end-to-end training. Such a heuristic approach lacks explicit statistical significance and controllability, often leading to a black-box optimization process that obscures the underlying structural correlations between sensor nodes. Furthermore, the joint optimization of adjacency matrices and network parameters can lead to topological forgetting or relational instability, thereby degrading the fidelity of topology awareness [19]. Consequently, constructing topological priors with rigorous statistical significance, interpretability, and transferability, while simultaneously enhancing continuous motion representation, remains a pivotal challenge in skeleton-based action recognition.

To address these challenges, this paper proposes KGS-GCN, a Graph Convolutional Network integrated with Kinematic Gaussian Splatting and Probabilistic Topology. Our framework re-envisions skeleton joint representation and topology construction through the lenses of kinematics and probability. By doing so, we effectively resolve the challenges of sensor data sparsity and topological rigidity inherent in conventional methods. Drawing inspiration from the success of 3D Gaussian Splatting in continuous radiance field representation and efficient rendering [24], the proposed method transforms discrete joints from deterministic points into probability distributions, explicitly encoding spatial uncertainty within the feature space. In contrast to computer graphics, which prioritizes static geometric representation, our approach emphasizes the kinematics-driven nature of sensor data. As joint velocity increases, the corresponding spatial distribution elongates anisotropically along the direction of motion. This mechanism ensures that velocity and orientation are naturally encoded as intrinsic features of the representation. Furthermore, formulating nodes as probability distributions enables the rigorous quantification of inter-node relationships via statistical distance. This establishes an interpretable prior for constructing semantic topologies that transcend physical connections. Specifically, the main contributions of this work are summarized as follows:

1. A kinematics-driven Gaussian splatting module is designed to dynamically construct anisotropic covariance matrices based on instantaneous joint velocities. This module renders sparse skeleton sequences into multi-view continuous heatmaps enriched with spatiotemporal semantics, significantly enhancing the representation of rapid motions and uncertainty.
2. KGS-GCN introduces a probabilistic topology construction strategy that quantifies statistical correlations via the Bhattacharyya distance between joint Gaussian distributions. This generates an interpretable adaptive prior adjacency matrix to complement physical topologies and capture latent long-range dependencies.
3. KGS-GCN incorporates a visual context gating mechanism that leverages rendered visual features to modulate feature propagation within the GCN backbone, facilitating the synergistic modeling of continuous visual representations and graph structure learning. Extensive experiments on multiple benchmark datasets validate the framework's efficacy in capturing complex spatiotemporal dynamics.

## II. Related Work

### A. Skeleton-Based Action Recognition

Skeleton-based action recognition aims to parse spatiotemporal dynamic patterns from sequences of human joint coordinates. Compared with RGB videos, skeleton representations are characterized by structural compactness and robustness against appearance perturbations, establishing them as a critical modality for action understanding. Early approaches primarily relied on handcrafted features or sequence models based on RNNs and CNNs, yet they faced inherent limitations in explicitly representing the non-Euclidean topology of the human body. ST-GCN [5] pioneered the modeling of skeleton sequences as spatiotemporal graphs, performing feature propagation along physical skeletal connections. This paradigm established the dominance of Graph Convolutional Networks (GCNs) in the field.

Following the ST-GCN paradigm, subsequent works have primarily evolved along three directions: adaptive topology learning, the design of efficient spatiotemporal operators, and the enhancement of topology awareness stability. Regarding adaptive topology, 2s-AGCN [6] introduced end-to-end learning of data-driven adjacency matrices. This mechanism allows graph structures to adaptively adjust according to specific samples or network layers, thereby overcoming the limitations of fixed physical connections. AS-GCN [7] and MS-AAGCN [9] reinforced spatiotemporal representation capabilities through structured relationship reasoning and multi-stream feature fusion, respectively. Furthermore, DGNN [8] utilized directed graph neural networks to mine asymmetric dependencies among skeletal joints. Targeting efficient modeling, Shift-GCN [10] replaced computationally intensive graph convolutions with lightweight spatial and temporal shift mechanisms, reducing complexity while expanding effective receptive fields. GCN-NAS [13] leveraged neural architecture search to automatically discover optimal network structures, aiming to enhance computational efficiency. Furthermore, recent works have focused on constructing robust and efficient baselines to reduce training and deployment costs [16].

To enhance operator expressiveness, existing works have focused on designing refined feature aggregation mechanisms. For instance, Disentangling GCN [11] proposed a disentangled graph convolution strategy to unify multi-scale feature aggregation and eliminate redundant dependencies. Similarly, Context-Aware GCN [12] incorporated a context-aware mechanism, enabling the network to effectively capture frame-wise global context information. Subsequently, CTR-GCN [14] refined topology modeling along the channel dimension to learn channel-specific relational structures. InfoGCN [15] introduced an information bottleneck objective to promote discriminative representation learning. Furthermore, DeGCN [20] integrated deformable sampling into spatial and temporal graph convolutions, adapting to intra-class deformations by learning dynamic receptive fields. Hierarchical decomposition and long-range connection modeling were employed to enhance multi-scale relational representations [17]. Meanwhile, Transformer architectures have further pushed performance boundaries by leveraging self-attention mechanisms to capture global long-range dependencies [21].

Despite continuous advancements in topology learning and spatiotemporal interaction modeling, mainstream frameworks typically treat joints as discrete point coordinates. This limitation hinders the explicit characterization of kinematic uncertainty in rapid motions and the associated spatial distribution along the velocity direction. Furthermore, although adaptive graph learning methods [6] capture latent long-range dependencies, the edge weights are typically learned implicitly as network parameters, resulting in a lack of explicit statistical significance and controllability. During joint optimization, the emphasis on topological information may be compromised or subject to forgetting, leading to the degradation of topology awareness [18-19]. Motivated by these observations, the proposed KGS-GCN formulates joints as probability distributions rather than deterministic points. By leveraging statistical distance to construct interpretable probabilistic topology priors, this approach provides a unified and interpretable perspective for continuous motion representation and topology learning.

### B. Gaussian Splatting

The field of neural rendering focuses on representing and rendering scenes in a differentiable manner. Neural Radiance Fields (NeRF) [22] achieved high-quality novel view synthesis via implicit neural fields, but they incur high training and inference costs. Meanwhile, related implicit neural representations employing periodic activation functions [23] have demonstrated strong fitting capabilities for continuous signals. In contrast, 3D Gaussian Splatting (3DGS) [24] adopts explicit Gaussian primitives to represent scenes, integrating differentiable rasterizers to achieve efficient rendering and high-quality reconstruction. This approach has significantly advanced the point-based explicit rendering paradigm. Regarding geometric consistency, 2D Gaussian Splatting (2DGS) [25] models object surfaces using surfel-based 2D Gaussian disks. This method improves geometric accuracy, establishing a solid foundation for subsequent extensions.

Benefiting from the efficiency of explicit representation and differentiable rendering, Gaussian splatting techniques have been rapidly applied to diverse tasks. In the realm of 2D image representation and compression, approaches such as GaussianImage [26] and Large Images Are Gaussians [27] utilize 2D Gaussian primitives to construct efficient image representations. Furthermore, Instant GaussianImage [28] explores image representation capabilities with enhanced generalization and adaptivity. Beyond generation and representation tasks, sparse Gaussian representations have been applied to data compression and knowledge distillation to enhance efficiency and scalability [29]. Furthermore, approaches like Speedy-Splat [30] accelerate the 3DGS pipeline through rendering and optimization improvements, while GaussianPro [31] refines training strategies via progressive propagation. For large-scale scene reconstruction, CityGaussian [32] improves training efficiency and real-time rendering capabilities in complex scenarios by employing divide-and-conquer and level-of-detail strategies. In the context of dynamic scene modeling, approaches such as Street Gaussians [33], MVSGaussian [34], and DrivingGaussian [35] incorporate temporal dimensions or dynamic attributes to capture moving objects and environmental variations. Additionally, Momentum-GS [36] reinforces the quality and stability of reconstruction through self-distillation and consistency constraints.

In general, existing Gaussian splatting approaches primarily target generative or reconstructive tasks, such as visual reconstruction, novel view synthesis, and compression. Even in dynamic settings, current works focus predominantly on improving the reconstruction quality of time-varying appearance and geometry [33-36]. In contrast, for discriminative tasks such as skeleton-based action recognition, systematic exploration is lacking regarding the transformation of anisotropic deformation features of Gaussian primitives into learnable kinematic cues and their synergistic modeling with graph structure learning. The proposed KGS-GCN introduces kinematics-driven anisotropic Gaussian splatting to explicitly encode joint velocity and directional information into spatiotemporal continuous joint heatmaps. Simultaneously, interpretable probabilistic topology priors are constructed based on the statistical distance between joint Gaussian distributions. This approach achieves the unified modeling of continuous motion representations and interpretable topology learning for skeleton-based action recognition.

## III. Proposed Method

This section details KGS-GCN, a graph convolutional network framework integrating kinematic Gaussian splatting and probabilistic topology. As illustrated in Fig. 1, the proposed method aims to mitigate the inherent limitations of discrete skeleton representations through kinematics-aware continuous

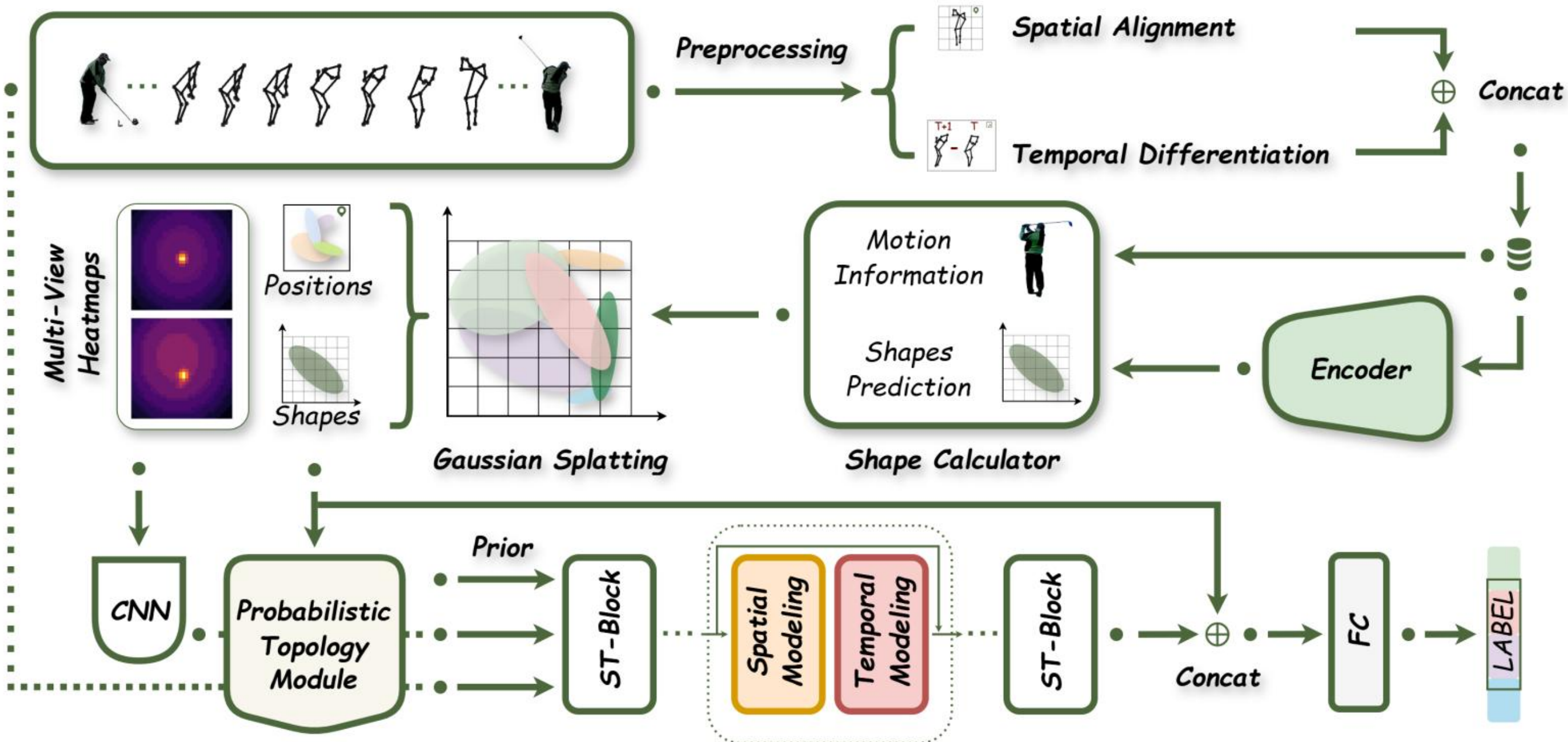

Fig. 1. The overall framework of KGS-GCN. Input skeleton sequences are processed to extract hybrid spatial and kinematic features. Anisotropic heatmaps and probabilistic joint distributions are generated via the Kinematics-driven Gaussian Splatting Module driven by hybrid features. Probabilistic topology is constructed utilizing statistical distance metrics to capture latent long-range dependencies. Discrete skeleton features from the backbone network and continuous visual cues from Gaussian maps are finally integrated to predict action classes.

**Algorithm 1** Overall Framework of KGS-GCN

**Input:** Skeleton Sequence $\mathbf{X} \in \mathbb{R}^{N \times C \times T \times V}$, One-hot Label $\mathbf{y}$

**Output:** Predicted Probability $\hat{\mathbf{y}}$

// 1. Kinematics-Driven Preprocessing
1: Compute Velocity: $\mathbf{V} \leftarrow \mathbf{X}_{t+1} - \mathbf{X}_t$;
2: Normalize $\mathbf{X}$ and $\mathbf{V}$ via SkeletonNorm to $[-1, 1]$;
// 2. Kinematic Gaussian Splatting (KGS)
3: Extract dynamic features: $\mathbf{F}_{dyn} \leftarrow \mathrm{Conv}(\mathrm{Concat}(\mathbf{X}, \mathbf{V}))$;
4: **for** *each joint $v$ at time $t$* **do**
5: Scale $\mathbf{S} \leftarrow \mathrm{BaseScale} \cdot (1 + \tanh(\|\mathbf{V}_{v,t}\|))$;
6: Rotation $\mathbf{R} \leftarrow \mathrm{Direction}(\mathbf{V}_{v,t})$;
7: $\mathbf{\Sigma}_{v,t} \leftarrow \mathbf{R}\mathbf{S}\mathbf{S}^T\mathbf{R}^T$;
8: **end**
9: Render Heatmaps $\mathbf{H}$ via Splatting using $\mathbf{X}$ (means) and $\mathbf{\Sigma}$;
// 3. Probabilistic Topology Construction
10: **for** *pair of joints $(i, j)$* **do**
11: Calculate Bhattacharyya Distance $D_B(\mathcal{N}_i, \mathcal{N}_j)$ using $\mathbf{X}$ and $\mathbf{\Sigma}$;
12: Compute Prior Adjacency: $\mathbf{A}_{prior}^{(i,j)} \leftarrow \exp(-D_B)$;
13: **end**
// 4. Visual Feature Encoding
14: Encode Heatmaps: $\mathbf{F}_{vis} \leftarrow \mathrm{VisualEncoder}(\mathbf{H})$;
// 5. GCN Backbone with Visual Gating
15: Initialize node features $\mathbf{Z}^{(0)} \leftarrow \mathrm{Embed}(\mathbf{X})$;
16: **for** layer $l = 1$ to $L$ **do**
17: Fuse Topology: $\mathbf{A}^{(l)} \leftarrow \mathbf{A}_{learn}^{(l)} + \beta \cdot \mathbf{A}_{prior}$;
18: Spatial Aggregation: $\mathbf{Z}_s^{(l)} \leftarrow \mathrm{GCN}(\mathbf{Z}^{(l-1)}, \mathbf{A}^{(l)})$;
// Visual Context Gating
19: Generate Gate: $\mathbf{G} \leftarrow \sigma(\mathrm{Project}(\mathbf{F}_{vis}))$;
20: Modulate: $\mathbf{Z}_{gate}^{(l)} \leftarrow \mathbf{Z}_s^{(l)} \odot (1 + \mathbf{G})$;
21: Temporal Aggregation: $\mathbf{Z}^{(l)} \leftarrow \mathrm{TCN}(\mathbf{Z}_{gate}^{(l)})$;
22: **end**
// 6. Prediction and Loss
23: Fusion: $\mathbf{F}_{final} \leftarrow \mathrm{Concat}(\mathrm{GAP}(\mathbf{Z}^{(L)}), \mathrm{GAP}(\mathbf{F}_{vis}))$;
24: Prediction: $\hat{\mathbf{y}} \leftarrow \mathrm{Softmax}(\mathrm{FC}(\mathbf{F}_{final}))$;
25: Compute Loss: $\mathcal{L} = \mathcal{L}_{CE}(\hat{\mathbf{y}}, \mathbf{y}) + \lambda \|\mathrm{sigmoid}(\mathbf{A}_{learn}) - \mathbf{A}_{prior}\|_2^2$;
26: **return** $\hat{\mathbf{y}}$

field modeling.

## A. Problem Definition and Overall Framework

The input sequence is represented as a five-dimensional tensor wherein each sample comprises $M$ pedestrians with $V$ joints spanning a sequence length $T$ with $C$ coordinate channels:

$$X \in \mathbb{R}^{N \times C \times T \times V \times M}, \quad y \in [1, K] \cap \mathbb{Z} \tag{1}$$

where N denotes the batch size, and K represents the number of classes. The KGS-GCN framework specifically targets two critical challenges: 1. the inadequacy of discrete, sparse joint coordinates in characterizing fine-grained motion blur and velocity; and 2. the restrictions of predefined skeleton topologies, which hinder the adaptive modeling of latent long-range dependencies.

Two core components are introduced to address this issue. Sparse skeletons are innovatively rendered into continuous multi-view heatmaps by the Kinematics-driven Gaussian Splatting Module (KGSM) alongside the explicit construction of velocity-driven anisotropic covariance. An innovative probabilistic topology construction strategy is concurrently proposed. Specifically, KGS-GCN formulates each joint as a Gaussian distribution and quantifies statistical correlations via the Bhattacharyya distance to generate a sample-adaptive prior adjacency matrix. Finally, the framework incorporates a visual context gating mechanism to inject the rendered visual context into the GCN backbone.

The detailed procedural steps of the entire framework are summarized in Algorithm 1 to elucidate the overall training procedure of KGS-GCN. Skeleton sequences are specifically transformed into multi-view heatmaps via kinematic computations and the Gaussian splatting module.

Algorithm 1 outlines the detailed training workflow of the proposed KGS-GCN. Initially, the framework transforms skeleton sequences into multi-view heatmaps by leveraging kinematic computations and the Gaussian Splatting Module. Subsequently, the probabilistic topology inferred via the Bhattacharyya distance initializes the adjacency matrix of the GCN. During feature aggregation, visual features modulate skeleton representations layer-by-layer through a gating mechanism. Finally, the model is updated by jointly minimizing the classification loss and the topology constraint loss.

## B. Kinematics-Driven Gaussian Splatting

Traditional sparse and discrete skeleton representations often overlook motion blur effects induced by velocity variations, which encapsulate rich temporal dynamic information. To address this limitation, the Kinematics-Driven Gaussian Splatting Module (KGSM) leverages instantaneous kinematic states to dynamically construct anisotropic Gaussian distributions.

Kinematic State Extraction and Normalization: To mitigate scale discrepancies across datasets, we implement an adaptive skeleton normalization strategy. Specifically, for a given frame $t$, we calculate the geometric center and translate the skeleton to the origin. Subsequently, we scale the coordinates to the interval $[-s, s]$ based on the maximum bounding sphere radius to obtain the normalized coordinates $\overline{X}$( $s$ is set to 0.8). Furthermore, to explicitly model motion trends, we compute the instantaneous velocity vector $V_{vel} \in \mathbb{R}^{T\times V\times C}$ for each joint:

$$v_{t,i} = \bar{x}_{t+1,i} - \bar{x}_{t,i} \tag{2}$$

where $v_{t,i}$ and $\bar{x}_{t,i}$ represent the velocity and position of the $i$ joint in the $t$ frame.

Dynamic Anisotropic Covariance Construction: This represents the core innovation of KGS-GCN. Unlike traditional Gaussian heatmaps that rely on fixed variances, we dynamically adapt the shape and orientation of Gaussian kernels according to joint velocities. Specifically, for each joint, we construct a 2D covariance matrix $\Sigma \in \mathbb{R}^{2\times 2}$ defined by the

$$\Sigma = RSS^{\top}R^{\top} \tag{3}$$

interplay of rotation $R$ and scaling matrices $S$, formulated as:

Regarding the construction of the scaling matrix $S = diag(s_x, s_y)$, we simulate motion blur by stretching the scale $s_x$ along the direction of motion as the velocity magnitude $\|v\|$ increases, while preserving the base scale $s_{\text{base}}$ in the perpendicular direction $s_y$. We formulate this process as:

$$s_x = s_{\text{base}} \cdot (1 + \alpha \tanh(\|v\|)),\ s_y = s_{\text{base}} \tag{4}$$

The stretching degree is controlled by the hyperparameter $\alpha$ set to 2 in this work whereas the adaptive baseline $s_{\text{base}}$ is learned by the network. The rotation matrix $R$ is determined by the direction of the velocity vector $(\cos\theta, \sin\theta)$ and is formulated as follows utilizing the normalized velocity direction :

$$R = \begin{bmatrix} \cos\theta & -\sin\theta \\ \sin\theta & \cos\theta \end{bmatrix} \tag{5}$$

According to the above definition, the elements of $\Sigma$ after expansion can be represented as:

$$\begin{cases} \Sigma_{xx} = s_x^2\cos^2\theta + s_y^2\sin^2\theta \\ \Sigma_{yy} = s_x^2\sin^2\theta + s_y^2\cos^2\theta \\ \Sigma_{xy} = (s_x^2 - s_y^2)\sin\theta\cos\theta \end{cases} \tag{6}$$

Through this formulation, stationary joints manifest as isotropic circular distributions, while rapidly moving joints appear as ellipses elongated along their motion trajectories. Consequently, this mechanism explicitly encodes temporal motion intensity directly within the spatial domain.

Implementation details regarding numerical stability: Numerical stability presents a significant challenge in dynamic covariance construction. This issue becomes particularly critical when joint velocities approach zero, as vanishing motion may theoretically lead to singular and non-invertible covariance matrices. To strictly ensure matrix invertibility and robust gradient propagation, we introduce a lower-bound truncation mechanism in the implementation. Specifically, the scale parameter $s$ is constrained above a strictly positive minimum threshold $\varsigma_{min}$ (set to $\varsigma_{min} = 0.1$ in this paper) before covariance construction. Consequently, the covariance of stationary joints safely degenerates into a stable isotropic Gaussian primitive, effectively preventing potential numerical collapse (e.g., division-by-zero errors) during the computation of the Bhattacharyya distance.

Multi-view Rendering: The 3D space is projected onto three orthogonal planes $(XY, YZ, ZX)$ to process 3D skeleton data wherein 2D Gaussian Splatting is independently executed on each view. The response intensity $G_{t,i}(p)$ generated by the $i$ joint at the $t$ frame follows a multivariate Gaussian distribution for an arbitrary pixel point $p \in \mathbb{R}^2$ on the plane:

$$G_{t,i}(p) = \exp\left(-\frac{1}{2}(p-\mu_{t,i})^{\top}\Sigma_{t,i}^{-1}(p-\mu_{t,i})\right) \tag{7}$$

where $\mu_{t,i} = x_{t,i}$ denotes the mean vector of the Gaussian distribution corresponding physically to the center position of the joint within the image coordinate system. The heatmap $H_t$ is formulated as the aggregation of responses from all joints for the final representation:

$$H_t(p) = \mathrm{A}\big((G_{t,i}(p))_{i=1}^{V}\big) \tag{8}$$

Sparse skeleton sequences are transformed into multi-channel continuous visual representations $F_{render} \in \mathbb{R}^{M\times T\times H\times W}$ via this process.

## C. Probabilistic Topology Construction

Latent dependencies among physically unconnected joints are frequently overlooked by physical connection graphs. We propose the construction of a probabilistic topology utilizing statistical parameters generated by KGSM to address this limitation. Each joint is modeled as a probability distribution $\mathrm{N}(\mu_i, \Sigma_i)$ wherein the correlation between joints $i$ and $j$ is quantified via the Bhattacharyya distance between their respective distributions. The analytical form is expressed as:

$$\begin{aligned} D_B(i,j) &= \frac{1}{8}\left(\mu_i - \mu_j\right)^T \Sigma_{avg}^{-1}\left(\mu_i - \mu_j\right) \\ &+ \frac{1}{2}\ln\left(\frac{\det\left(\Sigma_{avg}\right)}{\sqrt{\det\left(\Sigma_i\right)\det\left(\Sigma_j\right)}}\right) \end{aligned} \tag{9}$$

The first term of (9) quantifies the spatial Euclidean distance between joints whereas the second term measures the shape discrepancy between two distributions. $\Sigma_{avg}$ is expressed as:

$$\Sigma_{avg} = \frac{\Sigma_i + \Sigma_j}{2} \tag{10}$$

Based on the Bhattacharyya distance $D_B$, we constructed an adaptive prior adjacency matrix $A_{\text{prior}} \in \mathbb{R}^{V\times V}$:

$$A_{\text{prior}}(i,j) = \frac{1}{T}\sum_{t=1}^{T}\exp\left(-D_B(i_t, j_t)\right) \tag{11}$$

Long-range dependencies among joints based on motion statistical characteristics are adaptively captured by this matrix

$A_{prior}$. The matrix is subsequently injected into the following Graph Convolutional Network as prior knowledge.

### *D. Visual-Context Modulated GCN*

We construct the skeleton recognition network by stacking $n$ Spatio-Temporal Graph Convolutional Modules (ST-Blocks). As illustrated in Fig. 1, each ST-Block adopts the classic spatial-temporal decoupling design, consisting of two sequential sub-stages: the Visually-Enhanced Spatial GCN and the Multi-Scale Temporal TCN. The spatial modeling phase targets the capture of intra-frame joint dependencies. To this end, we employ a channel-level topology refinement mechanism and integrate visual context gating to achieve deep multi-modal feature fusion.

Initially, we process the rendered input $F_{\text{render}}$ via a lightweight CNN visual branch. This module incorporates multiple convolutional layers and downsampling operations to extract high-level visual semantic features. Subsequently, we apply global average pooling and linear projection to the output features to yield the visual context feature $F_{vis} \in \mathbb{R}^{N\times C^{'}\times T}$.

Visual Context Gating: This work designs a visual context gating mechanism within the GCN layer to achieve deep fusion of skeleton and visual features. Let $Y_{gcn} \in \mathbb{R}^{N\times C_{out}\times T\times V}$ denote the intermediate feature of the GCN layer. $F_{vis}$ is aligned and expanded to the identical dimension to generate modulation coefficients via a nonlinear gating network:

$$G = \sigma(W_g \cdot F_{vis} + b_g) \tag{12}$$

where $\sigma$ is the sigmoid activation function, $W_g$ is the learnable projection weights, and $b_g$is the learnable bias. The final fused features are calculated as follows:

$$\widetilde{Y} = Y_{gcn} \otimes (1 + G) + Y_{res} \tag{13}$$

where $\otimes$ denotes element-wise multiplication. Specific skeleton channel features are adaptively enhanced or suppressed according to the current action context via this residual gating mechanism to achieve the complementarity of multi-modal information.

Graph Convolution and Topology Fusion: The total adjacency matrix $A$ within each graph convolution layer is composed of the pre-defined physical graph $A_{phy}$, the network-learned graph $A_{\text{learn}}$ and the probabilistic topology $A_{\text{prior}}$ generated by our method. The feature aggregation process is formulated as:

$$Y_{gen} = \sum_{k} (A^{(k)}_{phy} + A^{(k)}_{learn} + \beta A_{prior}) X W_k \tag{14}$$

where $k$ is a learnable scaling factor used to dynamically adjust the importance of the prior probability topology.

Multi-Scale Temporal Convolution: This work designs a Multi-Scale Temporal Convolution Module (MS-TCN) along the temporal dimension to capture action patterns of varying durations. This module comprises multiple parallel convolution branches utilizing distinct dilation rates $d_k \in \{1,2,3,4\}$. The temporal output $Y_{out}$ is defined as the aggregation of outputs from respective branches for the fused feature $\widetilde{Y}$:

$$Y_{out} = \sum_{k} \text{Conv}_{d_k}(\text{ReLU}(\text{BN}(\text{Conv}_{1\times 1}(\widetilde{Y})))) + \widetilde{Y} \tag{15}$$

MS-TCN simultaneously captures short-term transient changes such as kicking instants and long-term action dependencies like walking periodicity via the combination of convolution kernels with diverse receptive fields. Unified modeling of complex spatiotemporal dynamics is thereby achieved.

Normalization and Integration of Probabilistic Topology: During training, we employ a learnable residual gating mechanism to normalize and integrate the constructed probabilistic topology. Rather than rigidly replacing the physical adjacency matrix, the normalized Gaussian topology serves as an adaptive spatial regularizer. Specifically, it is element-wise added to the dynamically learned spatial adjacency matrix, modulated by a learnable scaling parameter. This soft-gating paradigm seamlessly balances dataset-independent statistical structural priors with data-driven topological features, thereby yielding a stable and naturally normalized feature aggregation process.

### *E. Loss Function*

A multi-task loss function is adopted for model training. The standard cross-entropy loss $l_{ce}$ serves as the primary loss for the classification task. We introduce a topology consistency regularization term $l_{topo}$ to constrain the learned topology $A_{learn}$ within the GCN from deviating from statistical data regularities. The term is formulated as:

$$l_{topo} = \frac{1}{L}\sum_{l=1}^{L} \| \sigma(A^{(l)}_{\text{learn}}) - A_{\text{prior}} \|^2_F \tag{16}$$

where $L$ denotes the number of GCN layers whereas $\sigma$ represents the sigmoid activation function. $A_{\text{prior}}$ is treated as the pseudo label. The total loss function is formulated as:

$$l_{total} = l_{ce} + \lambda(t) \cdot l_{topo} \tag{17}$$

where $\lambda(t)$ denotes the weight coefficient dependent on epoch $t$. $\lambda(t)$ is initialized to a small value during the initial training phase to allow for free network exploration whereas the weight is gradually increased as training proceeds to enforce the constraining effect of the statistical prior.

## IV. Experiments

### *A. Experimental Setup*

*Datasets*: To comprehensively evaluate the effectiveness and generalization capabilities of the proposed KGS-GCN model, we conduct experiments on three widely used benchmark datasets: NTU RGB+D [43], NW-UCLA [45], and Penn Action [44]. Specifically, Penn Action and NW-UCLA focus on regular motion sequences, providing detailed human pose annotations and action categories. Thus, they serve as excellent testbeds for assessing the model's proficiency in modeling fine-grained motion details and handling posture variations. Conversely, NTU RGB+D, as one of the most extensively utilized large-scale 3D skeleton datasets, encompasses highly diverse action classes and scene variations. This enables a systematic examination of the model's robustness and cross-scenario generalization under complex conditions. Jointly evaluating these datasets facilitates a systematic and objective assessment of KGS-GCN across two critical dimensions: fine-grained action modeling capability and robustness in realistic, complex environments.

*Evaluation Metrics*: To align with standard comparison protocols in action recognition, we exclusively employ Top-1 Accuracy as the quantitative metric. This choice highlights the core performance of the model in terms of classification correctness. This metric measures the proportion of samples where the highest-probability prediction matches the ground truth. It directly indicates the model's recognition capability under standard classification settings, thereby facilitating consistent and fair comparisons across diverse experimental configurations.

TABLE I
QUANTITATIVE PERFORMANCE COMPARISON RESULTS ON DATASETS. RED AND BLUE INDICATE THE FIRST AND SECOND BEST RESULTS RESPECTIVELY FOR EACH INDIVIDUAL METRIC.

| Methods | NTU-60 (%) | | NTU-120 (%) | | Penn Action (%) | NW-UCLA (%) | Params (M) | Flops (G) |
|---|---|---|---|---|---|---|---|---|
| | x-sub | x-view | x-sub | x-set | | | | |
| MS-G3D [11] | 91.5 | 96.2 | 86.9 | 88.4 | 96.1 | - | 2.8 | 5.2 |
| CTR-GCN [14] | 92.4 | 96.4 | 88.9 | 90.4 | 96.9 | 96.5 | 1.5 | 2.0 |
| EfficientGCN [16] | 91.7 | 95.7 | 88.3 | 89.1 | 96.7 | - | 2.0 | 15.2 |
| InfoGCN [15] | 92.8 | 96.7 | 89.2 | 90.7 | 96.5 | 96.6 | 1.6 | 1.8 |
| FRHead [38] | 93.1 | 96.8 | 89.5 | 90.9 | 97.0 | 96.8 | 2.0 | - |
| BlockGCN [19] | 92.4 | 97.0 | 90.3 | 91.5 | 96.8 | 96.9 | **1.3** | **1.6** |
| DeGCN [20] | 93.3 | 97.4 | **91.0** | 92.1 | 97.6 | 97.2 | 5.6 | - |
| ST-TR [37] | 90.8 | 96.3 | 85.1 | 87.1 | 96.3 | - | 12.1 | 259.4 |
| TranSkeleton [39] | 92.8 | 97.0 | 89.4 | 90.5 | 96.7 | - | 2.2 | 9.2 |
| Hyperformer [40] | 92.9 | 96.5 | 89.9 | 91.3 | 97.1 | 96.7 | 2.7 | 9.6 |
| SkeMixFormer [41] | 93.0 | 97.1 | 90.1 | 91.3 | 99.2 | 97.4 | 2.1 | 4.8 |
| SkateFormer [21] | **93.5** | 97.4 | 89.8 | 91.4 | 98.4 | **98.3** | 2.0 | 3.6 |
| FreqMixFormer [42] | **93.6** | **97.4** | **90.5** | **91.9** | **99.7** | **97.4** | 2.0 | 64.4 |
| KGS-GCN | 92.8 | **97.2** | 88.9 | 90.8 | **99.5** | 97.3 | **1.4** | **1.3** |

TABLE II
QUANTITATIVE PERFORMANCE COMPARISON RESULTS OF CORE COMPONENT CONTRIBUTIONS. RED AND BLUE INDICATE THE FIRST AND SECOND BEST RESULTS RESPECTIVELY FOR EACH INDIVIDUAL METRIC.

| Method | +KGSM | +PT | +VCG | Penn Action (%) | NW-UCLA (%) |
|---|---|---|---|---|---|
| Baseline | × | × | × | 96.9 | 96.5 |
| Baseline+KGSM | √ | × | × | 98.0 | 96.7 |
| Baseline+KGSM+PT | √ | √ | × | **99.1** | 96.9 |
| Baseline+KGSM+VCG | √ | × | √ | 98.7 | **97.1** |
| KGS-GCN | √ | √ | √ | **99.5** | **97.3** |

*Implementation Details*: We implemented the KGS-GCN architecture based on the PyTorch framework. All training and inference phases were executed on a single NVIDIA RTX 4060 GPU. To optimize computational efficiency, we employed mixed-precision training. To ensure experimental reproducibility, we detail the parameter settings and training strategies as follows:

*Network Architecture Configuration*: We construct the KGS-GCN backbone using 10 stacked spatial-temporal graph convolution modules. The feature channels are set to 64, 128, and 256 for layers 1-4, 5-7, and 8-10, respectively. To expand the temporal receptive field and reduce computational overhead, we apply a temporal convolution stride of 2 in the 5th and 8th layers, while maintaining a stride of 1 elsewhere. Furthermore, we fix the rendered heatmap resolution at $32 \times 32$ and initialize the learnable log-scale parameter $log\,(\text{scale})$ to -2.0. To facilitate gradient backpropagation during the initial training phase, we configure the velocity stretching coefficient $\alpha$ to 2.0, ensuring small variance in the generated heatmaps. For the classification head, we project the encoded visual features into 128-dimensional vectors. Subsequently, we perform global average pooling and concatenate these visual vectors with the 256-dimensional skeleton features derived from the 10th layer, forwarding the fused representation to the fully connected layer.

*Training Strategy*: We train the network end-to-end using the SGD optimizer, with the Nesterov momentum set to 0.9 and weight decay fixed at $4 \times 10^{-4}$. We initialize the base learning rate at 0.05 and apply a multi-step decay schedule, scaling the rate by a factor of 0.1 at the 40th and 60th epochs. To prevent gradient instability during the early phase, we implement a linear warm-up strategy for the first 10 epochs, increasing the learning rate linearly from 0 to the base value. Regarding the loss function, we adopt a dynamic adjustment strategy, where $\lambda(t)$ is formulated as:

$$\lambda(t) = \lambda_{\text{base}} \times \min\,(1.0, t/5) \tag{18}$$

where $\lambda_{base} = 0.2$. This implies that topological constraints are gradually imposed during the initial training phase to provide a buffer period for adaptive network adjustment.

## B. Performance Comparison

We benchmark the proposed method against state-of-the-art approaches using the NTU [43], NW-UCLA [45], and Penn Action [44] datasets, with detailed comparisons provided in Table I. The experimental results highlight the significant advantages of our approach across all datasets. This performance is particularly noteworthy as we present the first framework to integrate Gaussian Splatting into this domain.

Specifically, KGS-GCN achieves competitive performance across various datasets while requiring only 1.4M parameters and 1.3 GFLOPs. On the NTU-60 benchmark, the model attains 92.8% accuracy on the x-sub split and 97.2% on the x-view split. This performance ranks second, trailing FreqMixFormer by a marginal gap of only 0.2%. Furthermore, on the NTU-120 x-sub and x-view benchmarks, KGS-GCN achieves accuracies of 88.9% and 90.8%, respectively, comparable to leading state-of-the-art methods. Given the substantial scale of the NTU-120 dataset and the lightweight nature of our model, these results validate the efficacy of the proposed Gaussian Splatting module and the probabilistic topology construction strategy.

TABLE III
ABLATION RESULTS OF SPLATTING STRATEGY ANALYSIS. RED INDICATES THE BEST RESULT FOR EACH INDIVIDUAL METRIC.

| Strategy | Covariance Type | Motion-Aware | Penn Action (%) |
|---|---|---|---|
| Isotropic Splatting | Scaled Identity | × | 97.9 |
| Kinematics-Driven | Anisotropic | √ | **99.5** |

TABLE IV
ABLATION RESULTS OF VISUAL FEATURE FUSION MECHANISM. RED INDICATES THE BEST RESULT FOR EACH INDIVIDUAL METRIC.

| Fusion Mechanism | Formula | Properties | Penn Action (%) |
|---|---|---|---|
| Element-wise Addition | $\widetilde{Y} = Y_{gcn} + F_{vis}$ | Equal Weight | 98.9 |
| Concatenation | $\widetilde{Y} = Concat(Y_{gcn}, F_{vis})$ | Channel Expansion | 99.2 |
| VCG | $\widetilde{Y} = Y_{gcn} \otimes (1+G) + Y_{res}$ | Adaptive Selection | **99.5** |

KGS-GCN exhibits solid generalization capabilities on smaller-scale datasets, such as NW-UCLA and Penn Action. Specifically, on the NW-UCLA benchmark, our model achieves comparable performance, showing a marginal difference of only 0.1% relative to the runner-up. Similarly, on the Penn Action dataset, KGS-GCN secures the second rank, trailing FreqMixFormer by a mere 0.2%.

In summary, KGS-GCN strikes an optimal balance between model complexity and inference performance, securing the second rank in parameter count and the top rank in computational efficiency. These results validate the efficacy of the proposed Gaussian Splatting module and probabilistic topology construction strategy, establishing a distinct competitive advantage over state-of-the-art methods.

## C. Ablation Study

To rigorously evaluate the contribution of each component within KGS-GCN, we conduct comprehensive ablation studies on the Penn Action and NW-UCLA datasets. We utilize CTR-GCN [14] as the baseline model for our backbone. Specifically, we examine the effectiveness of three key elements: the Kinematics-Driven Gaussian Splatting, the probabilistic topology construction strategy, and the visual context gating mechanism. To ensure fair comparisons, we maintain consistent training configurations across all experiments.

*Contribution of Individual Components*: We initially evaluate the impact of the core modules within KGS-GCN, specifically verifying the effectiveness of the Kinematics-Driven Gaussian Splatting Module (KGSM), the

TABLE V
ABLATION RESULTS OF METRIC FOR TOPOLOGY CONSTRUCTION. RED INDICATES THE BEST RESULT FOR EACH INDIVIDUAL METRIC.

| Metric | NW-UCLA (%) | Penn Action (%) |
|---|---|---|
| Euclidean Distance | 96.9 | 98.9 |
| Bhattacharyya | **97.3** | **99.5** |

Probabilistic Topology Construction strategy (PT), and the Visual Context Gating mechanism (VCG). As indicated in Table II, each component yields consistent performance improvements. Most notably, the complete framework outperforms the baseline network by 2.6% on the Penn Action dataset and 0.8% on NW-UCLA. These outcomes empirically validate the efficacy of our architectural design.

*Splatting Strategy Analysis*: A core motivation of KGS-GCN lies in explicitly modeling motion blur effects via anisotropic covariance matrices. To validate this design, we compare the proposed kinematics-driven strategy against a standard isotropic counterpart. The isotropic approach disregards velocity vectors, limiting the covariance matrix to a scaled identity matrix. As indicated in Table III, our kinematics-driven method surpasses the isotropic baseline by 1.6%, empirically confirming the efficacy of the proposed strategy.

*Mechanism of Visual Feature Fusion:* To identify the optimal strategy for integrating visual context into the GCN backbone, we compare three distinct fusion mechanisms: Element-wise Addition, Concatenation, and our proposed Visual Context Gating (VCG). As presented in Table IV, simple addition and concatenation yield only marginal performance improvements. We attribute this limitation to potential background noise and spatial redundancy in the rendered heatmaps, which can corrupt high-level skeleton semantics during direct fusion. In contrast, the VCG mechanism effectively modulates the feature flow, outperforming the addition and concatenation strategies by 0.6% and 0.3%, respectively. Consequently, these results validate VCG as the superior choice for our framework.

*Metric for Topology Construction*: We examine the metrics used to construct probabilistic topology priors. Specifically, we benchmark the Bhattacharyya distance adopted in this work against the conventional Euclidean distance derived from joint coordinates. As shown in Table V, the Bhattacharyya distance yields superior performance, validating its selection for KGS-GCN.

*Robustness Under Degraded Sensing Conditions:* To validate the effectiveness of our framework under sparse sensing conditions, we first conduct a systematic robustness evaluation on the Penn Action dataset across three categories of sensor degradation. Specifically, we simulate: 1) spatial sparsity by stochastically dropping joints with probabilities of 20%, 40%, and 60%; 2) temporal sparsity by uniformly downsampling the input sequences by factors of 2× and 4×; and 3) sensor noise by injecting Gaussian perturbations with standard deviations of 5, 15, and 25mm into the joint coordinates. All compared methods are trained exclusively on pristine (clean) data and subsequently evaluated under these identical degradation settings to assess their zero-shot robustness.

As illustrated in Table VI, the experimental results yield several pivotal findings. First, KGS-GCN consistently outperforms all baseline counterparts across the entire spectrum of degradation settings; notably, the performance margin widens significantly as the severity of degradation intensifies. Under heavy joint dropout ($p = 50\%$), KGS-GCN surpasses SkateFormer by 3.7%, a marked increase compared to its 1.1% lead under pristine conditions. Second, KGS-GCN demonstrates exceptional resilience to temporal downsampling, retaining 97.9% accuracy even at a 4× downsampling factor, whereas all baseline methods fall significantly below 93.7%. Finally, under severe coordinate jitter ( $\sigma = 25mm$ ), KGS-GCN sustains a robust accuracy of 96.2%, while the strongest baseline degrades to 91.6%. These results substantiate the tangible advantages of KGS-GCN in practical sparse and noisy sensing environments, demonstrating a utility that transcends standard benchmark performance.

TABLE VI
ABLATION RESULTS OF ROBUSTNESS UNDER DEGRADED SENSING CONDITIONS ON THE PENN ACTION BENCHMARK. RED INDICATES THE BEST RESULT FOR EACH INDIVIDUAL METRIC.

| Degradation Type | Level | CTR-GCN [14] | InfoGCN [15] | BlockGCN [19] | SkateFormer [21] | KGS-GCN |
|---|---|---|---|---|---|---|
| Clean | - | 96.9 | 96.5 | 96.8 | **98.4** | **99.5** |
| Joint Dropout | $p$=20% | 92.4 | 94.3 | 94.6 | **95.6** | **97.7** |
| | $p$=40% | 87.2 | 89.7 | 90.4 | **91.7** | **95.8** |
| | $p$=50% | 81.9 | 84.2 | 88.7 | **89.9** | **93.6** |
| Temporal Downsampling | $r$=2× | 94.3 | 95.0 | 95.3 | **96.4** | **99.1** |
| | $r$=4× | 91.3 | 91.7 | 92.1 | **93.7** | **97.9** |
| Gaussian Noise | $\sigma$=5$mm$ | 95.7 | 95.7 | 96.2 | **97.2** | **99.3** |
| | $\sigma$=15$mm$ | 92.4 | 92.5 | 93.7 | **95.4** | **98.4** |
| | $\sigma$=25$mm$ | 87.3 | 88.3 | 89.7 | **91.6** | **96.2** |

TABLE VII
ABLATION RESULTS OF ROBUSTNESS UNDER DEGRADED SENSING CONDITIONS ON THE NTU RGB+D 60 X-SUB BENCHMARK. RED INDICATES THE BEST RESULT FOR EACH INDIVIDUAL METRIC.

| Degradation Type | Level | CTR-GCN [14] | InfoGCN [15] | BlockGCN [19] | SkateFormer [43] | KGS-GCN |
|---|---|---|---|---|---|---|
| Clean | - | 92.4 | **92.8** | 92.4 | **93.5** | **92.8** |
| Joint Dropout | $p$=20% | 85.1 | 86.3 | 87.0 | **88.2** | **88.7** |
| | $p$=40% | 78.8 | 80.9 | 82.5 | **84.1** | **85.7** |
| | $p$=50% | 72.2 | 74.5 | 77.6 | **79.3** | **84.2** |
| Temporal Downsampling | $r$=2× | 90.7 | 91.2 | 91.5 | **92.3** | **92.1** |
| | $r$=4× | 85.5 | 86.1 | 87.2 | **88.1** | **89.8** |
| Gaussian Noise | $\sigma$=5$mm$ | 91.2 | 92.5 | 92.6 | **93.1** | **93.3** |
| | $\sigma$=15$mm$ | 86.4 | 86.9 | 88.3 | **89.9** | **90.1** |
| | $\sigma$=25$mm$ | 79.6 | 80.8 | 83.3 | **85.2** | **87.9** |

TABLE VIII
ABLATION RESULTS OF IMPACT OF COVARIANCE CLAMPING THRESHOLD ON NUMERICAL STABILITY. RED INDICATES THE BEST RESULT FOR EACH INDIVIDUAL METRIC.

| $\varsigma_{min}$ | Penn Action (%) |
|---|---|
| 0.01 | 98.7 |
| 0.05 | **99.2** |
| 0.10 (Ours) | **99.5** |
| 0.20 | 99.1 |
| 0.50 | 98.4 |

To verify that the robustness gains generalize from fine-grained data to large-scale complex scenarios, we reproduce the same degradation protocol on NTU RGB+D 60 X-sub. Given the evaluation cost of combining three degradation types with multiple severity levels, we select X-sub as the representative subset because it is the most widely adopted and already covers diverse subjects and scenes. As shown in Table VII, the trends observed on Penn Action hold here as well: KGS-GCN performs slightly below SkateFormer on clean data, yet surpasses all baselines once degradation is introduced, with the margin increasing as degradation intensifies. Under joint dropout $p = 50\%$, KGS-GCN achieves 84.2%, leading SkateFormer by 4.9%; under coordinate jitter $\sigma = 25\ mm$, it still maintains 87.9%. These results corroborate the robustness of KGS-GCN under realistic low-fidelity sensing conditions.

Impact of Covariance Clamping Threshold on Numerical Stability. To explicitly address potential numerical instabilities such as covariance matrix singularity arising from negligible motion displacement, KGS-GCN incorporates a lower-bound truncation threshold $\varsigma_{min}$ for the Gaussian variance parameters. We empirically investigate the sensitivity of recognition performance to this stability threshold, with results summarized in Table VIII. Our findings indicate that an insufficiently small $\varsigma_{min}$ occasionally precipitates numerical underflow during matrix inversion, which hinders model convergence and leads to suboptimal results. Conversely, an excessively large $\varsigma_{min}$ results in over-smoothed probability distributions, thereby obscuring fine-grained kinematic nuances and diminishing the discriminative power of the topological structure. Experimental results demonstrate that a moderate $\varsigma_{min}$ strikes an optimal balance: it rigorously ensures numerical robustness across all network layers while preserving the model's intrinsic sensitivity to subtle spatiotemporal dynamics.

## V. Conclusion

We propose KGS-GCN, a graph convolutional network that integrates kinematics-driven Gaussian Splatting with probabilistic topology. By conceptualizing skeleton data as generative sources of continuous visual signals, our approach significantly enhances the capacity to model complex spatiotemporal dynamics. Ultimately, this framework provides a novel perspective for the unified modeling of skeleton and visual features. Our core contribution lies in the proposal of the Kinematics-driven Gaussian Splatting Module. By leveraging instantaneous joint velocities, this module dynamically constructs anisotropic covariance matrices. This mechanism effectively recovers the motion blur and spatiotemporal continuity lost in discrete coordinates, thereby providing semantic inputs significantly richer than raw positional information. Furthermore, we challenge the convention of fixed or attention-based topologies by introducing a probabilistic topology construction strategy. By modeling joints as Gaussian distributions and quantifying their overlap via the Bhattacharyya distance, we derive a graph structure that captures intrinsic statistical motion correlations. This formulation proves robust against noise arising from non-physically connected joints. Ultimately, by employing rendered visual context to modulate geometric graph convolutions, our unified architecture establishes a novel synergy between low-level kinematics and high-level visual semantics. While these results are encouraging, there remains potential for further investigation. In future work, we aim to develop lightweight approximation methods for probabilistic topology to enhance computational efficiency. Additionally, we intend to explore the application of KGS-GCN in generative tasks, where continuous Gaussian representations can provide superior smoothness and enhanced interpretability.

## References

[1] W. Xin, R. Liu, Y. Liu, *et al*., "Transformer for skeleton-based action recognition: A review of recent advances," *Neurocomputing*, vol. 537, pp. 164–186, 2023.

[2] R. Yue, Z. Tian, and S. Du, "Action recognition based on RGB and skeleton data sets: A survey," *Neurocomputing*, vol. 512, pp. 287–306, 2022.

[3] Y. Kong and Y. Fu, "Human action recognition and prediction: A survey," *Int. J. Comput. Vis*., vol. 130, no. 5, pp. 1366–1401, 2022.

[4] J. Zhang, L. Lin, S. Yang, *et al*., "Self-supervised skeleton-based action representation learning: A benchmark and beyond," *arXiv preprint* arXiv:2406.02978, 2024.

[5] S. Yan, Y. Xiong, and D. Lin, "Spatial temporal graph convolutional networks for skeleton-based action recognition," in *Proc. AAAI Conf. Artif. Intell.*, vol. 32, no. 1, 2018.

[6] L. Shi, Y. Zhang, J. Cheng, *et al*., "Two-stream adaptive graph convolutional networks for skeleton-based action recognition," in *Proc. IEEE/CVF Conf. Comput. Vis. Pattern Recognit. (CVPR)*, 2019, pp. 12026–12035.

[7] M. Li, S. Chen, X. Chen, *et al*., "Actional-structural graph convolutional networks for skeleton-based action recognition," in *Proc. IEEE/CVF Conf. Comput. Vis. Pattern Recognit. (CVPR)*, 2019, pp. 3595–3603.

[8] L. Shi, Y. Zhang, J. Cheng, *et al*., "Skeleton-based action recognition with directed graph neural networks," in *Proc. IEEE/CVF Conf. Comput. Vis. Pattern Recognit. (CVPR)*, 2019, pp. 7912–7921.

[9] L. Shi, Y. Zhang, J. Cheng, *et al*., "Skeleton-based action recognition with multi-stream adaptive graph convolutional networks," *IEEE Trans. Image Process*., vol. 29, pp. 9532–9545, 2020.

[10] K. Cheng, Y. Zhang, X. He, *et al*., "Skeleton-based action recognition with shift graph convolutional network," in *Proc. IEEE/CVF Conf. Comput. Vis. Pattern Recognit. (CVPR)*, 2020, pp. 183–192.

[11] Z. Liu, H. Zhang, Z. Chen, *et al*., "Disentangling and unifying graph convolutions for skeleton-based action recognition," in *Proc. IEEE/CVF Conf. Comput. Vis. Pattern Recognit. (CVPR)*, 2020, pp. 143–152.

[12] X. Zhang, C. Xu, and D. Tao, "Context aware graph convolution for skeleton-based action recognition," in *Proc. IEEE/CVF Conf. Comput. Vis. Pattern Recognit. (CVPR)*, 2020, pp. 14333–14342.

[13] W. Peng, X. Hong, H. Chen, *et al*., "Learning graph convolutional network for skeleton-based human action recognition by neural searching," in *Proc. AAAI Conf. Artif. Intell*., vol. 34, no. 3, 2020, pp. 2669–2676.

[14] Y. Chen, Z. Zhang, C. Yuan, *et al*., "Channel-wise topology refinement graph convolution for skeleton-based action recognition," in *Proc. IEEE/CVF Int. Conf. Comput. Vis. (ICCV)*, 2021, pp. 13359–13368.

[15] H. Chi, M. H. Ha, S. Chi, *et al*., "InfoGCN: Representation learning for human skeleton-based action recognition," in *Proc. IEEE/CVF Conf. Comput. Vis. Pattern Recognit. (CVPR)*, 2022, pp. 20186–20196.

[16] Y.-F. Song, Z. Zhang, C. Shan, *et al*., "Constructing stronger and faster baselines for skeleton-based action recognition," *IEEE Trans. Pattern Anal. Mach. Intell*., vol. 45, no. 2, pp. 1474–1488, 2022.

[17] J. Lee, M. Lee, D. Lee, *et al*., "Hierarchically decomposed graph convolutional networks for skeleton-based action recognition," in *Proc. IEEE/CVF Int. Conf. Comput. Vis. (ICCV)*, 2023, pp. 10444–10453.

[18] Y. Zhou, Z.-Q. Cheng, J.-Y. He, *et al*., "Overcoming topology agnosticism: Enhancing skeleton-based action recognition through redefined skeletal topology awareness," *arXiv preprint* arXiv:2305.11468, 2023.

[19] Y. Zhou, X. Yan, Z.-Q. Cheng, *et al*., "BlockGCN: Redefine topology awareness for skeleton-based action recognition," in *Proc. IEEE/CVF Conf. Comput. Vis. Pattern Recognit. (CVPR)*, 2024, pp. 2049–2058.

[20] W. Myung, N. Su, J.-H. Xue, *et al*., "DE-GCN: Deformable graph convolutional networks for skeleton-based action recognition," *IEEE Trans. Image Process*., vol. 33, pp. 2477–2490, 2024.

[21] J. Do and M. Kim, "Skateformer: Skeletal-temporal transformer for human action recognition," in *Proc. Eur. Conf. Comput. Vis. (ECCV)*, 2024, pp. 401–420.

[22] B. Mildenhall, P. P. Srinivasan, M. Tancik, J. T. Barron, R. Ramamoorthi, and R. Ng, "NeRF: Representing scenes as neural radiance fields for view synthesis," *Commun. ACM*, vol. 65, no. 1, pp. 99–106, 2021.

[23] V. Sitzmann, J. Martel, A. Bergman, D. Lindell, and G. Wetzstein, "Implicit neural representations with periodic activation functions," in *Adv. Neural Inf. Process. Syst. (NeurIPS)*, vol. 33, 2020, pp. 7462–7473.

[24] B. Kerbl, G. Kopanas, T. Leimkühler, and G. Drettakis, "3D Gaussian splatting for real-time radiance field rendering," *ACM Trans. Graph.*, vol. 42, no. 4, Art. no. 139, 2023.

[25] B. Huang, Z. Yu, A. Chen, A. Geiger, and S. Gao, "2D Gaussian splatting for geometrically accurate radiance fields," in *ACM SIGGRAPH Conf. Papers*, 2024, pp. 1–11.

[26] X. Zhang, X. Ge, T. Xu, *et al*., "GaussianImage: 1000 FPS image representation and compression by 2D Gaussian splatting," in *Proc. Eur. Conf. Comput. Vis. (ECCV)*, 2024, pp. 327–345.

[27] L. Zhu, G. Lin, J. Chen, *et al*., "Large images are Gaussians: High-quality large image representation with levels of 2D Gaussian splatting," in *Proc. AAAI Conf. Artif. Intell. (AAAI)*, 2025, pp. 10977–10985.

[28] Z. Zeng, Y. Wang, C. Yang, T. Guan, and L. Ju, "Instant GaussianImage: A generalizable and self-adaptive image representation via 2D Gaussian splatting," *arXiv preprint* arXiv:2506.23479, 2025.

[29] C. Jiang, Z. Li, H. Zhao, Q. Shan, S. Wu, and J. Su, "Beyond pixels: Efficient dataset distillation via sparse Gaussian representation," *arXiv preprint* arXiv:2509.26219, 2025.

[30] A. Hanson, A. Tu, G. Lin, V. Singla, M. Zwicker, and T. Goldstein, "Speedy-Splat: Fast 3D Gaussian splatting with sparse pixels and sparse primitives," in *Proc. IEEE/CVF Conf. Comput. Vis. Pattern Recognit. (CVPR)*, 2025, pp. 21537–21546.

[31] K. Cheng, X. Long, K. Yang, Y. Yao, W. Yin, Y. Ma, *et al*., "GaussianPro: 3D Gaussian splatting with progressive propagation," in *Proc. Int. Conf. Mach. Learn. (ICML)*, 2024.

[32] Y. Liu, C. Luo, L. Fan, N. Wang, J. Peng, and Z. Zhang, "CityGaussian: Real-time high-quality large-scale scene rendering with Gaussians," in *Proc. Eur. Conf. Comput. Vis. (ECCV)*, 2024, pp. 265–282.

[33] Y. Yan, H. Lin, C. Zhou, W. Wang, H. Sun, K. Zhan, *et al.*, "Street Gaussians: Modeling dynamic urban scenes with Gaussian splatting," in *Proc. Eur. Conf. Comput. Vis. (ECCV)*, 2024, pp. 156–173.

[34] T. Liu, G. Wang, S. Hu, L. Shen, X. Ye, Y. Zang, *et al.*, "MVSGaussian: Fast generalizable Gaussian splatting reconstruction from multi-view stereo," in *Proc. Eur. Conf. Comput. Vis. (ECCV)*, 2024, pp. 37–53.

[35] X. Zhou, Z. Lin, X. Shan, Y. Wang, D. Sun, and M.-H. Yang, "DrivingGaussian: Composite Gaussian splatting for surrounding dynamic autonomous driving scenes," in *Proc. IEEE/CVF Conf. Comput. Vis. Pattern Recognit. (CVPR)*, 2024, pp. 21634–21643.

[36] J. Fan, W. Li, Y. Han, T. Dai, and Y. Tang, "Momentum-GS: Momentum Gaussian self-distillation for high-quality large scene reconstruction," in *Proc. IEEE/CVF Int. Conf. Comput. Vis. (ICCV)*, 2025, pp. 25250–25260.

[37] C. Plizzari, M. Cannici, and M. Matteucci, "Skeleton-based action recognition via spatial and temporal transformer networks," *Comput. Vis. Image Understand.*, vol. 208, Art. no. 103219, 2021.

[38] H. Zhou, Q. Liu, and Y. Wang, "Learning discriminative representations for skeleton based action recognition," in *Proc. IEEE/CVF Conf. Comput. Vis. Pattern Recognit. (CVPR)*, 2023, pp. 10608–10617.

[39] H. Liu, Y. Liu, Y. Chen, C. Yuan, B. Li, and W. Hu, "TransSkeleton: Hierarchical spatial–temporal transformer for skeleton-based action recognition," *IEEE Trans. Circuits Syst. Video Technol.*, vol. 33, no. 8, pp. 4137–4148, 2023.

[40] Y. Zhou, Z.-Q. Cheng, C. Li, Y. Fang, Y. Geng, X. Xie, and M. Keuper, "Hypergraph transformer for skeleton-based action recognition," *arXiv preprint* arXiv:2211.09590, 2022.

[41] W. Xin, Q. Miao, Y. Liu, R. Liu, C.-M. Pun, and C. Shi, "Skeleton MixFormer: Multivariate topology representation for skeleton-based action recognition," in *Proc. ACM Int. Conf. Multimedia (ACM MM)*, 2023, pp. 2211–2220.

[42] W. Wu, C. Zheng, Z. Yang, C. Chen, S. Das, and A. Lu, "Frequency guidance matters: Skeletal action recognition by frequency-aware mixed transformer," in *Proc. ACM Int. Conf. Multimedia (ACM MM)*, 2024, pp. 4660–4669.

[43] A. Shahroudy, J. Liu, T.-T. Ng, *et al.*, "NTU RGB+D: A large scale dataset for 3D human activity analysis," in *Proc. IEEE Conf. Comput. Vis. Pattern Recognit. (CVPR)*, 2016, pp. 1010–1019.

[44] W. Zhang, M. Zhu, and K. G. Derpanis, "From actemes to action: A strongly-supervised representation for detailed action understanding," in *Proc. IEEE Int. Conf. Comput. Vis. (ICCV)*, 2013, pp. 2248–2255.

[45] J. Wang, X. Nie, Y. Xia, *et al.*, "Cross-view action modeling, learning and recognition," in *Proc. IEEE Conf. Comput. Vis. Pattern Recognit. (CVPR)*, 2014, pp. 2649–2656.

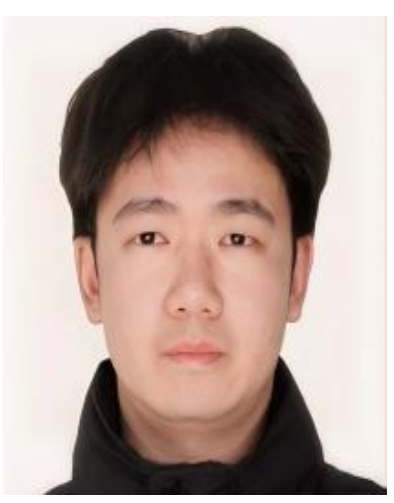

**Yuhan Chen** received his master's degree in 2024 from the College of Mechanical Engineering at Chongqing University of Technology. He is currently pursuing the Ph.D. degree in College of Mechanical and Vehicle Engineering at Chongqing University, China. His research interests include deep learning, Low-level Vision and Gaussian Splatting.

.

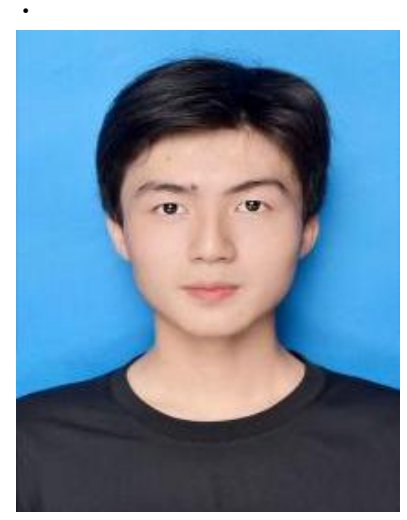

**Yicui Shi** received the B.E degree majoring in Automotive Engineering at Chongqing University in 2025. He is currently pursuing the M.S. degree in Automotive Engineering at Chongqing University, Chongqing, China. His research interests include computer vision and Gaussian Splatting.

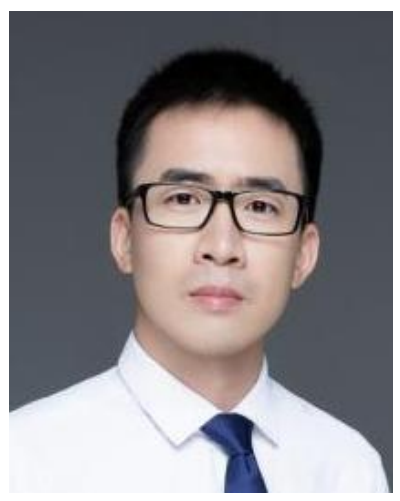

**Guofa Li** received the Ph.D. degree in Mechanical Engineering from Tsinghua University, China, in 2016. He is currently a Professor with Chongqing University, China. His research interests include environment perception, driver behavior analysis, and smart decision-making based on artificial intelligence technologies in autonomous vehicles and intelligent transportation systems. He serves as the Associate Editor for *IEEE Transactions on Intelligent Transportation Systems, IEEE Transactions on Affective Computing, and IEEE Sensors Journal*.

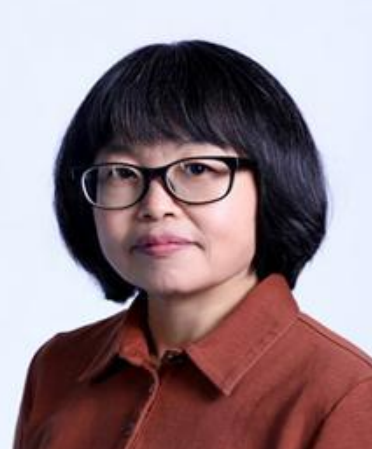

**Liping Zhang** is currently a tenured Professor in Department of Mathematical Sciences, Tsinghua University. She received her Ph.D. degree from the Academy of Mathematics and Systems Sciences, Chinese Academy of Sciences in 2001. Her research interests include continuous optimization, tensor analysis and computation, machine learning. She has published more than 70 research papers in international journals such as *Mathematical Programming, SIAM Journal on Optimization, Mathematics of Computation, Mathematics of Operational Research, SIAM Journal on Matrix Analysis and Applications, Journal of Machine Learning Research, Expert Sysytem with Applications,* etc.

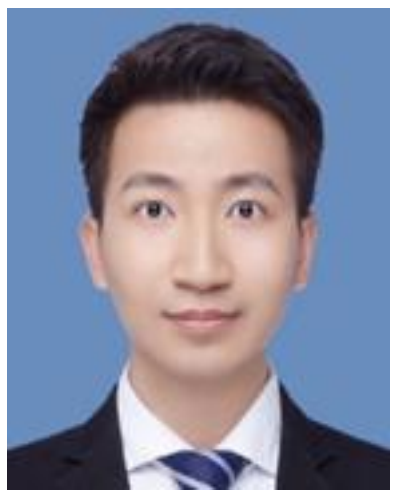

**Jie Li** received the Ph.D. degree in mechanical engineering from Tsinghua University, Beijing, China, 2024. He is currently an Associate Professor with the College of Mechanical and Vehicle Engineering, Chongqing University, Chongqing, China. His research interests include model predictive control, adaptive dynamic programming and reinforcement learning.

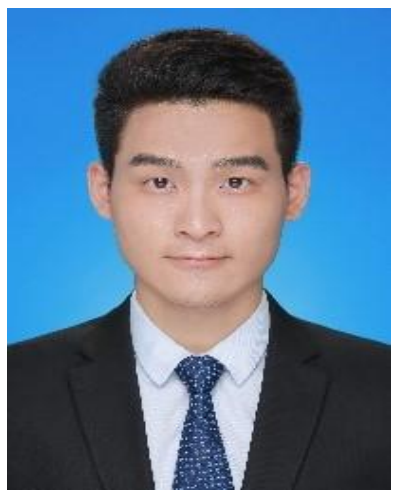

**Jiaxin Gao** received his B.S. and Ph.D. degrees from the University of Science & Technology Beijing in 2017 and 2023, respectively. He is currently an Assistant Researcher at the School of Vehicle and Mobility, Tsinghua University. His current research interests focus on reinforcement learning, decision and control for autonomous vehicles, and perceptual data generation for autonomous driving.

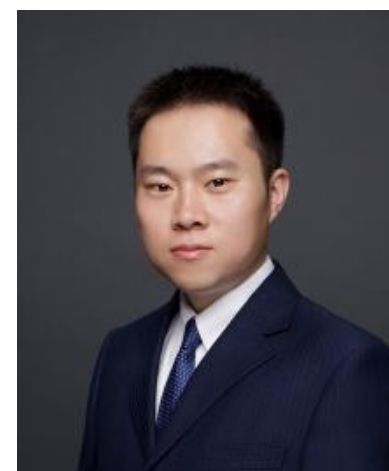

**Wenbo Chu** received the B.S. degree in automotive engineering from Tsinghua University, Beijing, China, in 2008, the M.S. degree in automotive engineering from RWTH Aachen, Germany, and the Ph.D. degree in mechanical engineering from Tsinghua University, in 2014.

He is currently a research fellow with the Western China Science City Innovation Center of Intelligent and Connected Vehicles (Chongqing) Co., Ltd., and National Innovation Center of Intelligent and Connected Vehicles. He is also the doctoral supervisor with Chongqing University, and the professor with Chongqing University of Technology. His research interests include intelligent and connected vehicles, new energy vehicles, and vehicle dynamics and control.